\documentclass[preprint,12pt]{elsarticle}
\usepackage{graphicx}
\usepackage[margin=1.0in]{geometry}
\usepackage{color, colortbl}
\usepackage{hyperref}
\usepackage{float}
\usepackage{amsmath}
\usepackage{caption}
\usepackage{subcaption}
\usepackage[normalem]{ulem}

\definecolor{LightCyan}{rgb}{0.88,1,1}
\definecolor{LightRose}{rgb}{1,0.88,0.88}
\definecolor{LightGreen}{rgb}{0.88,1,0.88}

\title{Using Machine Learning for Particle Track \\
Identification in the CLAS12 Detector}
\author[1]{Polykarpos Thomadakis \fnref{fn1,fn2}}
\author[1]{Angelos Angelopoulos \fnref{fn1}}
\author[2]{Gagik Gavalian \fnref{fn1}}
\author[1]{Nikos Chrisochoides}

\address[1]{CRTC, Department of Computer Science, Old Dominion University, Norfolk, VA, USA}
\address[2]{Jefferson Lab, Newport News, VA, USA}

\fntext[fn1]{Authors contributed equally.}
\fntext[fn2]{Correspoding author \textit{pthom001@odu.edu}}

\begin{document}


\begin{abstract}
Particle track reconstruction is the most computationally intensive process in nuclear physics experiments. Traditional algorithms use a combinatorial approach that exhaustively tests track measurements (``hits'') to identify those that form an actual particle trajectory. In this article, we describe the development of four machine learning (ML) models that assist the tracking algorithm by identifying valid track candidates from the measurements in drift chambers. Several types of machine learning models were tested, including: Convolutional Neural Networks (CNN), Multi-Layer Perceptrons (MLP), Extremely Randomized Trees (ERT) and Recurrent Neural Networks (RNN). As a result of this work, an MLP network classifier was implemented as part of the CLAS12 reconstruction software to provide the tracking code with recommended track candidates. The resulting software achieved accuracy of greater than 99\% and resulted in an end-to-end speedup of 35\% compared to existing algorithms.
\end{abstract}
\maketitle
\section{Introduction}
\indent

In nuclear physics, experiments measuring scattered particle parameters are the most computationally-intensive process. This process relies on measurements of particle tracking detectors to construct a particle trajectory by combining the detected hits and resolving the particle momentum via fitting the trajectory points (using Kalman Filter\cite{Kalman1960}). In high luminosity experiments (where multiple particles are produced as a result of an interaction, and noise is present in particle tracking detectors), the process of isolating detector hits for each particle trajectory relies on considering each combination of hits that can potentially form a track and then fitting each hypothesis to determine which one represents a valid trajectory. This process can be time-consuming, amounting to about $94\%$ of the total data post-processing time.

\paragraph{Motivation}
Recent advances in artificial intelligence and machine learning create the opportunity  for substituting  some of the existing algorithms  with predictions from machine learning models. 
 With this substitution, we reduce the complexity of the code needed to select the correct track hit combinations by providing only the most likely  track trajectory candidates.  
In this work, we focus on the track-candidate identification process for the CLAS12\cite{Burkert:2020akg} detector at Jefferson Laboratory (JLab), Newport News, Virginia. We study different types of machine learning models, including CNN, ERT, MLP, and RNN. The goal of our investigation is to construct a model that can identify the candidate with the highest probability of representing a real track out of all possible track candidates formed by the combination of different particle detections.
We evaluate each model for its accuracy and speed of inference to determine their effectiveness on CLAS12's tracking performance. The resulting implementation produces a speedup of $35\%$ in the tracking code and a track candidate identification accuracy of $99.9\%$ when compared to tracks reconstructed by the current algorithm.

\section{CLAS12 Detector}
\indent

The CLAS12\cite{Burkert:2020akg} detector is built around a six-coil toroidal magnet which divides the active detection into six azimuthal regions, called ``sectors''. The torus coils are approximately planar. Each sector subtends an azimuthal range of 60$^\circ$ from the mid-plane of one coil to the mid-plane of the adjacent coil. The “sector mid-plane” is an imaginary plane that bisects the sector’s azimuth.

\begin{figure}[!ht]
\begin{center}
 \includegraphics[width=3.5in]{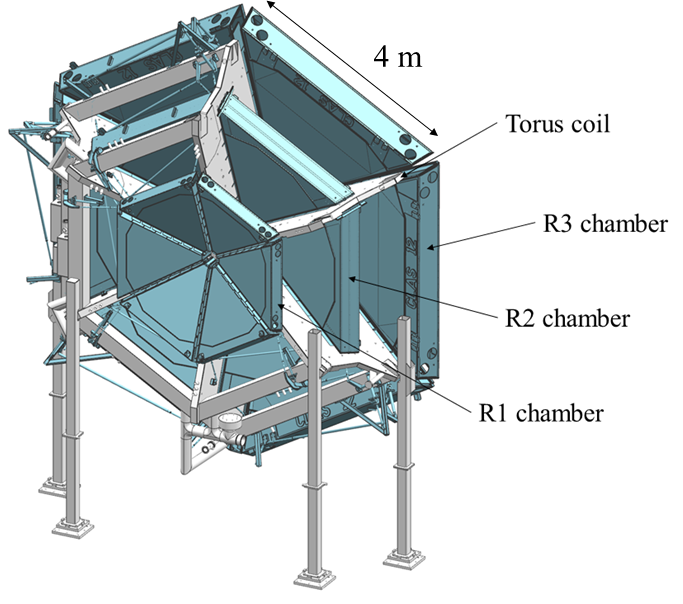}
\caption {CLAS12 Drift Chambers inside the toroidal field (figure reproduced from~\cite{Burkert:2020akg}, coauthored by Dr. Gavalian). Three different Regions (R1, R2 and R3) consist of two Super Layers of chambers with 6 layers of wires in the Super Layer.}
 \label{dc:geometry}
 \end{center}
\end{figure}

\begin{figure}[!ht]
\begin{center}
 \includegraphics[width=3.5in]{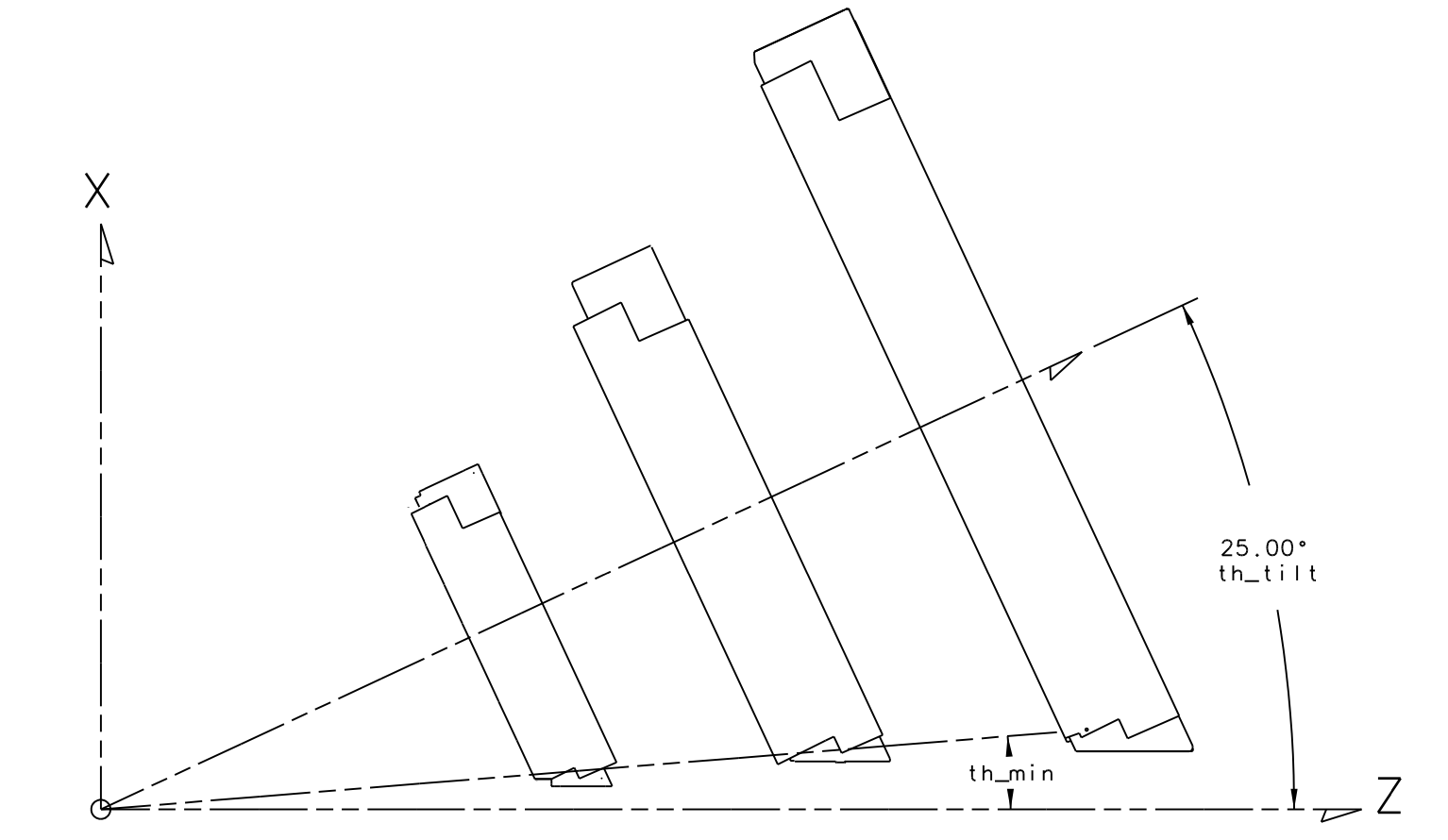}
 \includegraphics[width=2.5in]{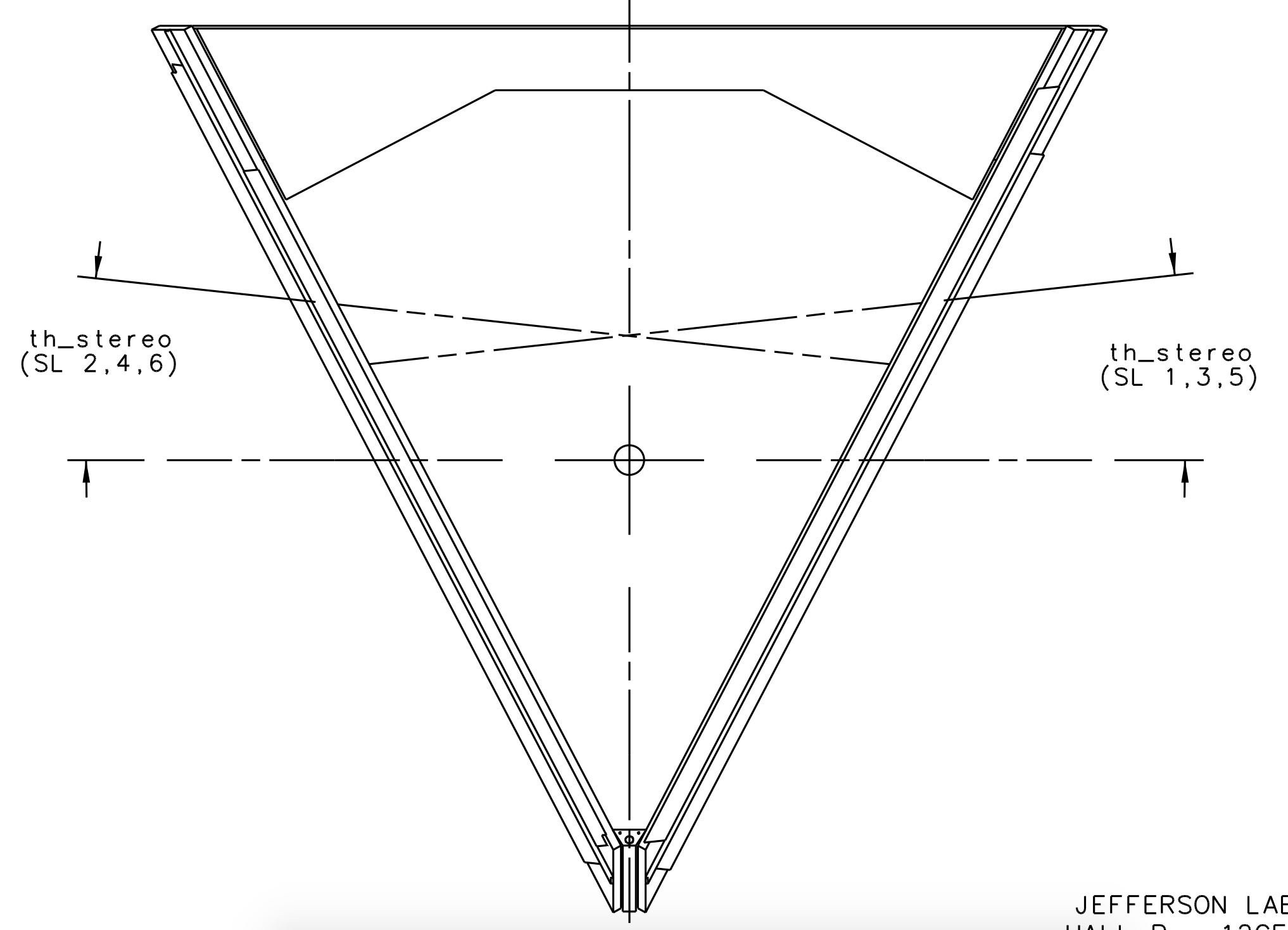}
\caption {Side and front view of drift chambers (figure reproduced from~\cite{Mestayer:2020saf}, coauthored by Dr. Gavalian). a) the layout for one sector showing three regions of drift chamber. Z-axis is direction of incoming beam. b) diagram of wire directions in each super layer.}
 \label{dc:geometrysketch}
 \end{center}
\end{figure}

Charged particles in the CLAS12 detector are tracked using drift chambers\cite{Mestayer:2020saf} inside the toroidal magnetic field (Figure \ref{dc:geometry}). There are six identical independent drift chamber systems in CLAS12 (one for each azimuthal sector); each of them consists of three chambers (called ``regions'') separated from each other along the beam direction, as shown in Figure~\ref{dc:geometrysketch}a. Each region  consists of two parts called super-layers with wires in each super-layer running at 6$^\circ$ relative to the main axis of the chambers, shown in Figure~\ref{dc:geometrysketch}b. Particles passing through the drift chambers leave a signal in the wires.

\paragraph{Reconstruction Procedure}

The reconstruction software uses clusters of hits to compose a trajectory through the drift chambers. First, neighboring hits in each super-layer of the drift chambers are grouped together to form clusters. Then, track candidates are constructed from all combinations of six clusters in one drift chamber sector. Each track candidate is fitted using a polynomial function to determine the particle's initial momentum and origin angles.

\begin{figure}[!ht]
\begin{center}
\includegraphics[width=6.5in]{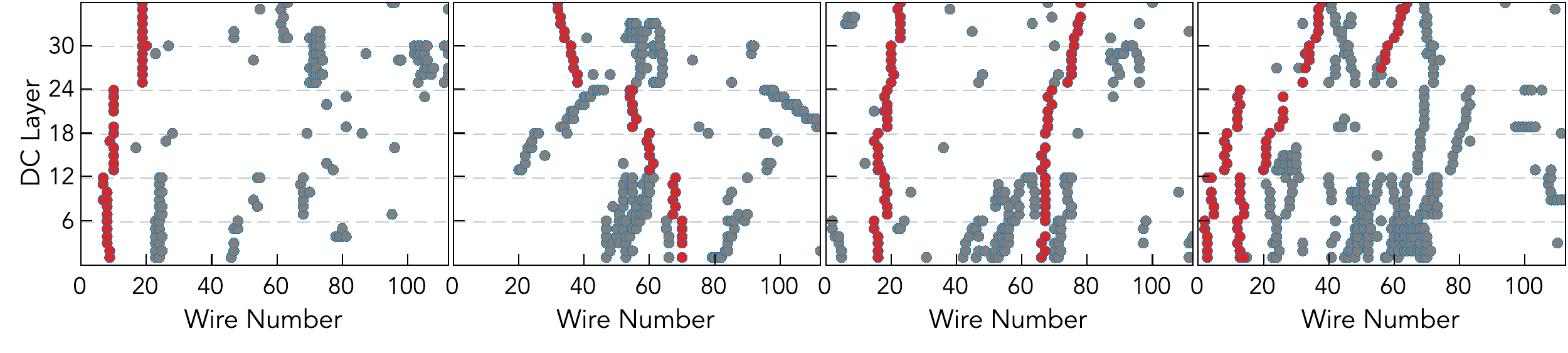}
\caption { Example of Drift Chamber data with wire hits (gray circles) and hits identified as belonging to a track by tracking algorithm (red circles). Each plot presents data from one sector from different events. Cases with one and two tracks in one sector are shown. }
 \label{event:view}
 \end{center}
\end{figure}

Track candidates that satisfy the ``good fit'' requirement are further processed (fitted) using Kalman Filter which outputs the final track parameters, such as momentum, polar and azimuthal angles at the vertex.
Examples of reconstructed tracks (for one sector) are shown in Figure~\ref{event:view} where all hits in drift chambers are shown with gray points, and hits belonging to a reconstructed track are shown with red points. Each reconstructed track has an associated cluster in each of the super-layers ( super-layer boundaries indicated by dashed lines), and plots with 12 clusters are events  in which two tracks were reconstructed . As can be seen in the example plots, there are many more clusters  that are not part of a track (background clusters, shown in gray), and  their number is even higher in two-track events.
All cluster combinations have to be considered by the tracking algorithm before detecting the valid track candidates and running them through the final Kalman Filter. Thus, as the number of clusters increases, so does the cost of the process (e.g., time, money). By using ML, we  determine which candidates represent a possible good track and have the tracking algorithm analyze only those. This may result in tracking code speed-up and possibly lead to much simpler and maintainable code for track candidate selection. For training our model, we use data that was processed by the conventional tracking algorithm and good tracks were already isolated (similar to the data shown in Figure~\ref{event:view}).

\section{Related Work}
Farrel et al.~\cite{Farrell:2018cjr} used space-point data arranged as sequences and connected graphs.
For the first approach an RNN is presented with a sequence of hit coordinates and predicts the 
coordinates of the next hits. The authors experimented with this approach using a traditional 
regression model as well as a Gaussian distribution model for the produced predictions. Their results
on low occupancy tests showed very good performance for both models. For the second approach, they presented the data as a graph of connected hits and use a Graph Neural Network (GNN) to perform track reconstruction. Two paths were examined, one using GNNs to classify graph 
nodes as part of the same track, while the other performs classification on graph edges instead and is
able to detect many tracks at once.

In another effort, Farrel et al.~\cite{Farrell:2017} investigated the applicability of image-based and
point-based models. In the image-based approach, they presented each layer of the detector as an image
which is fed to an RNN that predicts the correct location of a target track. A variation of this approach
used a CNN that is fed with the entire detector image and classifies the points belonging to the same
track. While this approach performed well, the authors argued that it would not be possible to scale it up
as the dimensionality and sparsity increases. In the point-based approach, the spacepoints of the
detector are sorted before being fed to an RNN which can, then, classify the spacepoints to the 
respective target tracks. The results presented in this work were collected by running on toy data, not real experiments. Tsaris et al.~\cite{tsaris2018hep} employed the models developed here
to more realistic datasets, and showed that they can achieve promising results.

In \cite{baranov2017novel}, Baranov et al. proposed a two-step technique consisting of a hit preprocessing
step, followed by the employment of a deep neural network that classifies detected hits as parts of a
particle track. In the preprocessing step, the authors use a
directed search algorithm to filter possible candidates and narrow the search space. 
The deep neural network that follows consists of a 
one-dimensional convolutional layer followed by an RNN with two Gated Recurrent Unit (GRU) layers. The filtered hits are
fed to the neural network which is able to classify them as valid points and assign them to their respective track. With this approach, they achieve
performance of up to 97.5\% for a dataset created using a Monte Carlo generator. An improved version
of this model, in terms of accuracy and speed, was presented in \cite{baranov2019particle} where the authors replaced the two-step approach with a single neural network. 
This network enhanced the neural network presented in the original work with a regression component that estimated the location of a hit in the next detector layer.

\section{Data Selection and Performance Metrics}

\subsection{Data Selection}

Due to inefficiencies that develop over time, not all of the six layers in a given
super-layer have hits in each of the clusters. For these reasons, we used different data representation (features) for each of the models.
For the CNN, an image of size 36x112 was used, representing all the wires in one sector of the drift chamber, since the inefficiencies get smoothed out
by convolutional and pooling layers. For the ERT and MLP models, we used six-feature input for each track, representing the average wire number of the cluster
in each super-layer. For the RNN, we pre-processed the data to produce a sequence of 36 numbers that represent the track trajectory.
For this purpose, each cluster in all six super-layers was fitted with a linear function to determine  its intercept and slope. Then, the signals in missing layers were completed with a pseudo hit with the wire number that lies on the line representing the cluster. After this procedure we had a track candidate with 36 input features, one for each wire in the drift chambers layers.

To generate the training datasets, we extract good and bad tracks from data events that have already been processed by the conventional algorithm and write out their hit patterns. An example of one event is shown in Figure~\ref{event:view}, where all hit clusters are presented along with the clusters that form a valid track. 


In all representations, the cluster combinations that form a real track  are labeled as a positive sample, while  those that did not get identified as valid tracks are labeled as negative samples. Since there are multiple negative samples per event alongside the positive one, we need to balance the training dataset between negative and positive samples. To achieve this, we only provide a single negative sample for each positive sample of an event. In the next section, we evaluate how different options for negative samples affect our results.

\subsection{Training Data Tuning}
\label{sec:training_data_opt}

When selecting training data, we discovered that negative sample selection can affect the model accuracy.
In each reconstructed event with a valid track, there are many track candidates that we have to choose as a
false sample. In the training dataset, we balance the number of ``true'' and ``false'' samples in order to prevent the machine learning model from overfitting to ``false cases''.

\begin{figure}[!ht]
\begin{center}
\includegraphics[width=6in]{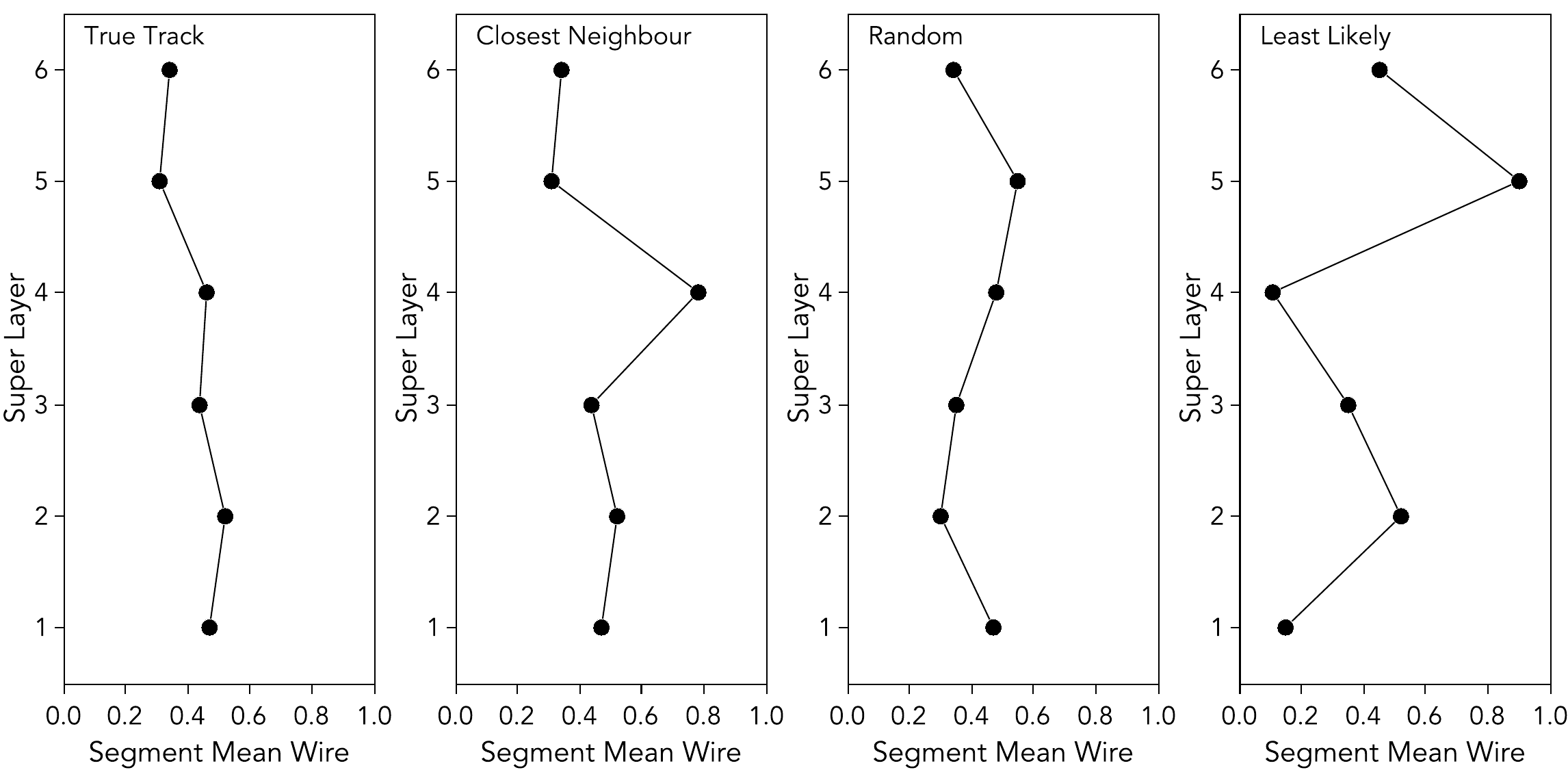}
\caption {Training data selection for the inference accuracy study. Examples of data samples for ``Least likely'',
``Random'' and ``Closest Neighbor''. (Wire numbers are normalized to 112)}
 \label{ai::training_sample}
 \end{center}
\end{figure}

We create three training datasets (shown on Firgure~\ref{ai::training_sample}) and test the network
trained on different input data. The samples and choices for the ``false'' track candidate shown on
Figure~\ref{ai::training_sample} are the following:

\begin{itemize}
\item track candidate that looks least like the real track in that event 
\item track candidate randomly chosen from all combinations  
\item track candidate that is closest to the real track; most of the time they differ only by one cluster 
\end{itemize}

Three models are trained using three datasets, and then model evaluation is done on a subset of
``closest neighbor'' data.

\begin{figure}[!ht]
\begin{center}
 \includegraphics[width=2.1in]{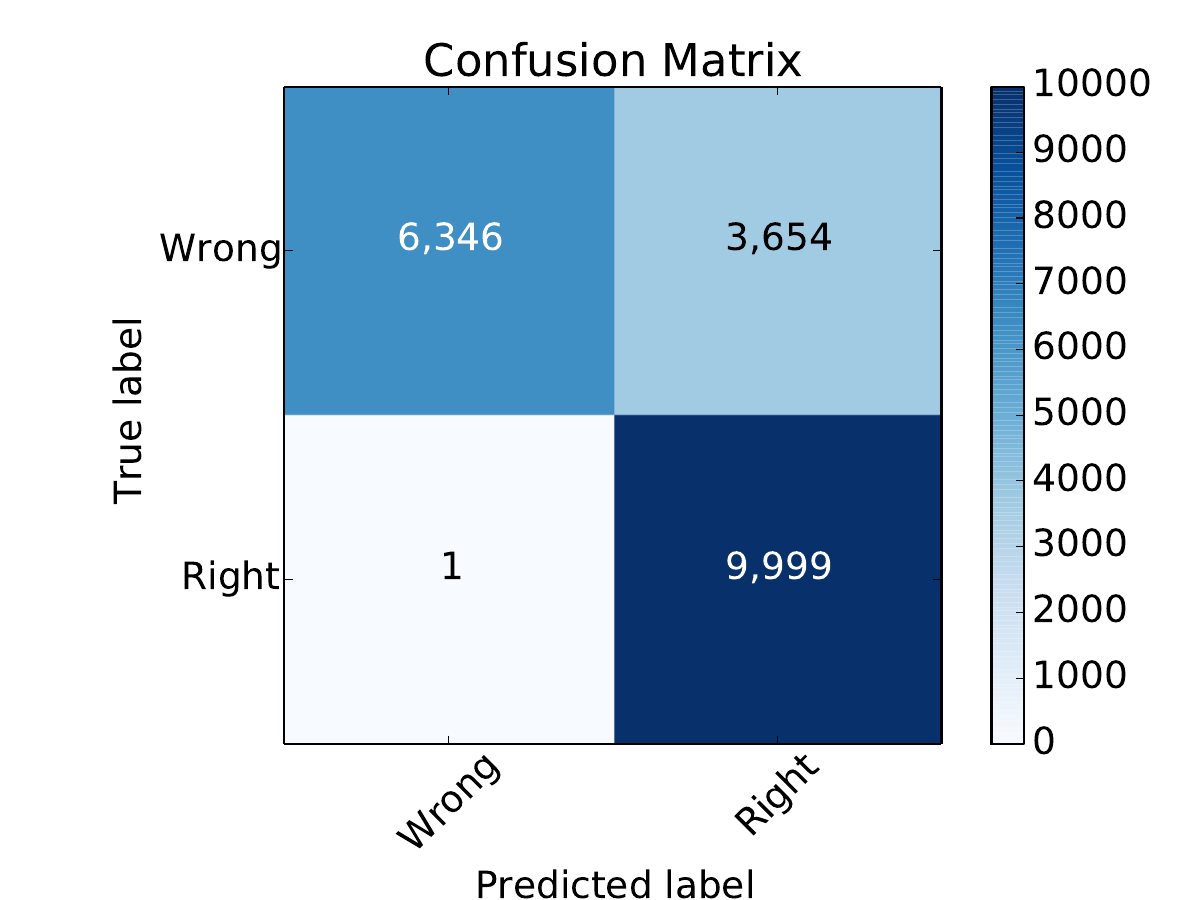}
 \includegraphics[width=2.1in]{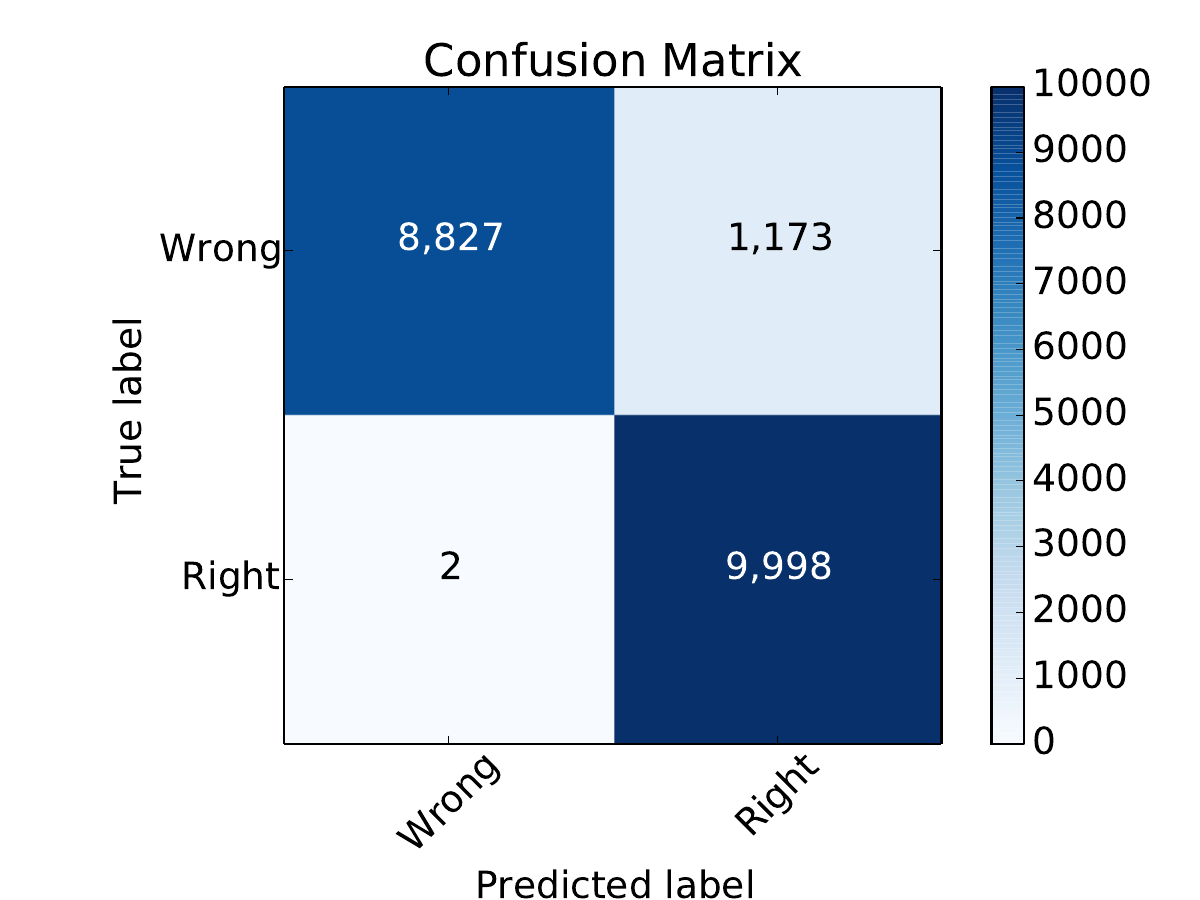}
 \includegraphics[width=2.1in]{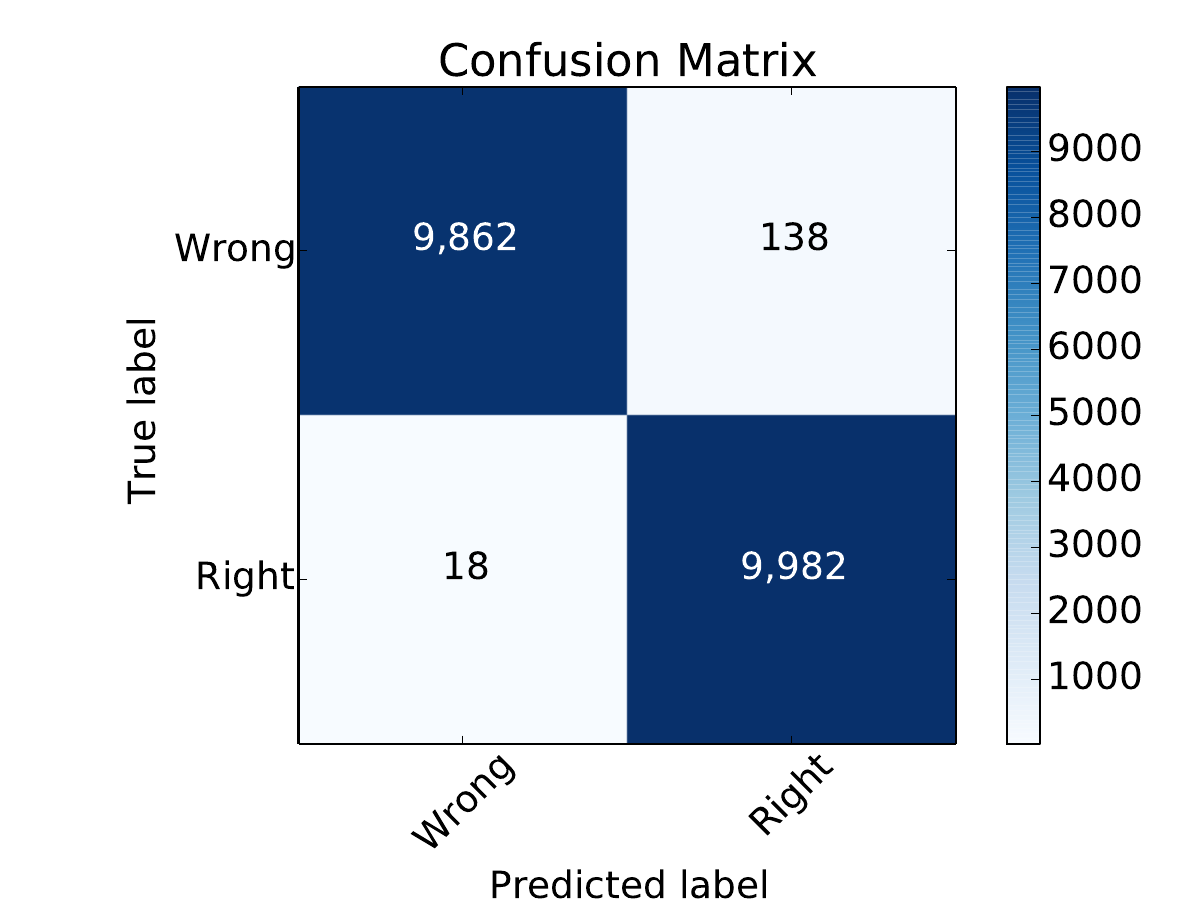}
\caption {Confusion matrix (number of events) for different models : (left) trained with data set with least likely track candidates, (middle) trained with sample with randomly chosen ``false'' candidates, (right) trained with sample closest neighbor ``false'' candidates. }
 \label{ai::confusion_matrix_tests}
 \end{center}
\end{figure}

As can be seen from the confusion matrices in Figure~\ref{ai::confusion_matrix_tests}, the model trained on random and ``least likely'' datasets is able to accurately predict the ``true'' track but fails in classifying the ``false'' track.  Because the model has not been shown a false candidate very similar to a true candidate, it is not able to distinguish false candidates very similar to the true. The tests show that model accuracy depends on training sample composition. The best results are achieved using the ``closest neighbor'' training dataset. This method is used for all model evaluations and for the final implementation of training data extraction from experimental data.

\subsection{Performance Metrics}

In order to determine our models' accuracy, we devised and utilized several metrics in addition to using the standard accuracy metrics. The need for these new metrics  stems from the fact that the performance of the machine learning models should be evaluated in the context of event samples rather than the overall performance on the whole dataset. Moreover, it is crucial to identify the valid track in each sample; however, it is  less harmful to misclassify an invalid track since it will be eliminated by the following stage of the tracking algorithm where Kalman-Filter fitting is applied. Of course, it is important to keep false positive number low as well since Kalman-Filter fitting is the slowest part of the reconstruction process. 
Also, examining only the accuracy of the model does not give a good indication about its overall performance due to the heavily imbalanced amount of invalid tracks in the testing sample. For example, a model that classifies all tracks as invalid may achieve relatively high accuracy;  however, it is in fact useless, because it cannot detect valid tracks. 

Our custom metrics provide a better indication about the real performance of the machine learning methods for the needs of our application. The accuracy metrics consist of:
\begin{enumerate}
	\item \textbf{A-det}: The ratio of samples where the valid particle track was correctly detected.
	\item \textbf{A-det$|$conf}: The percentage of A-det for which there were invalid tracks confused (misidentified) as valid ones (false positives).
	\item \textbf{A-det$|$high}: The percentage of A-det for which the valid particle track had the highest probability of being valid out of all tracks in a sample.
	\item \textbf{A-notdet}: The ratio of samples where the valid track was not detected (false negatives). This metric was very critical for us to minimize, as we don't want to miss valid particle tracks.
\end{enumerate}

\section{Models Description and Performance Evaluation}
\subsection{Evaluation Settings}
The following evaluations were performed on Old Dominion University's Wahab High-Performance Computing Cluster. 
The training set consists of 3.4 million tracks spread equally between valid and invalid tracks. The evaluation dataset consists of 14,760 events totalling 606,223 tracks, both generated from real data in CLAS12 after being classified using the conventional algorithm. Clasifications for models outputting probabilities (ERT, MLP, CNN) are made around a threshold of 0.5; tracks with inferred probability higher or equal to that are considered valid. The ERT and MLP models executed on a 40-core Intel Xeon Gold 6148 CPU @ 2.40GHz, utilizing the 6-feature data format and were implemented using the scikit-learn library\cite{scikitlearn}. The  CNN and RNN executed on a NVIDIA Tesla V100 GPU. The CNN utilized the 36x112-feature data format while RNN the 36-feature format. Both networks were implemented in TensorFlow 2 \cite{tensorflow} with Python 3 on a Linux Ubuntu machine.   
\subsection{Extremely Randomized Trees}

\begin{figure}
    \centering
    \includegraphics[width=0.7\textwidth]{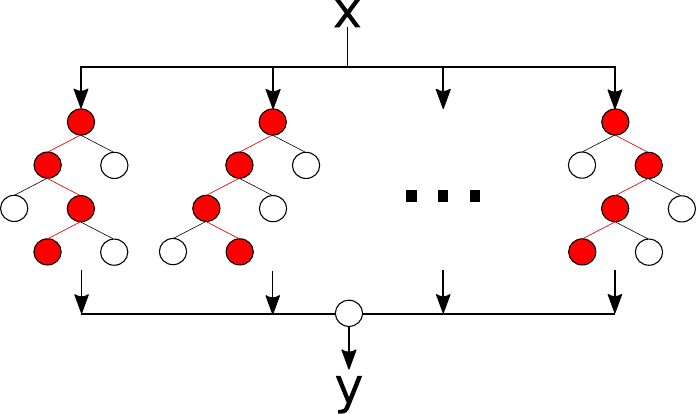}
    \caption{An illustration of the extremely randomized trees decision-making algorithm}
    \label{fig:ert_example}
\end{figure}

ERT~\cite{ERT-Paper} is a supervised learning algorithm that constitutes an ensemble of randomly generated decision trees \cite{decisionTreesQuinlan}. A decision tree is formed by recursively dividing the dataset into subsets. The splitting criteria are formed by the algorithm based on the classification features and such that the derived subsets are optimal. The recursive splitting process completes when the derived subset consists only of same class elements or when splitting does not add any extra value. However, this algorithm introduces a high probability of overfitting the training data, in other words, creating a model that cannot generalize to new data. For example, it could generate a tree with a leaf for each example in the training set which would not be reusable when used on a different dataset. To mitigate such issues, one can use multiple decision trees formed by random subsets of the dataset. The randomness introduced by generating new decision trees in this way dramatically improves the prediction power of the model. A method that follows this approach by picking random subsets of the input dataset with replacement is called random forest \cite{ho1995random}. ERT is an extension of the random forest method that incorporates more randomness by randomly selecting the splitting criteria instead of choosing the best split. Moreover,  this method forms the different random trees by using subsets of the input dataset without replacement. The final prediction of this model is produced by taking the average of the predictions of all random trees. Figure \ref{fig:ert_example} shows an example of the ERT decision algorithm.

\paragraph{Architecture} 
For our application, we use a model with three hundred estimators (decision trees), with no limit in the number of features to be considered when splitting. In addition, we use the information gain (entropy) split quality criterion. The rest of the parameters are kept at their default values. The input for the model is a 6-feature dataset.

\paragraph{Results}
The ERT model has 99.96\% accuracy in identifying the valid particle track in a sample (A-det metric). In  40.72\% of the samples where the valid track is identified, there are also some false positives (A-det$|$conf metric); however, in most cases ($>97\%$ of the samples), the valid track is given the highest probability of being valid  (A-det$|$high metric). Finally, there are almost no samples where the valid track is not detected (A-notdet metric). These results are compiled in Table \ref{tab: ert-metrics}. Table \ref{tab: ert-confmatrix} shows the confusion matrix generated by  evaluating the model on the testing dataset, which indicates very few cases of false negatives and an acceptable amount of false positives.

\begin{table}[ht]
    \centering
    \begin{tabular}{|l|l|}
    \hline
    \textbf{Metric}   & \textbf {Result}  \\ \hline
    A-det                & 99.96\% \\ \hline
    A-det$|$conf                & 40.72\%    \\ \hline
    A-det$|$high                & 97.09\%  \\ \hline
    A-notdet                & 0.04\%   \\ \hline
    Time to Train     & 502 sec   \\ \hline
    Time to Predict/sample & 306 $\mu s$ \\ \hline
    \end{tabular}
    \caption{Shows the results for some metrics used to evaluate the model. The ERT model executed on a multi-core CPU.}
    \label{tab: ert-metrics}
\end{table}

\begin{table}[ht]
    \centering
    \begin{tabular}{|l|l|l|}
    \hline
    & \textbf{Predicted: Invalid} & \textbf{Predicted: Valid} \\ \hline
    \textbf{Actual: Invalid} & 560,019              & 31,444             \\ \hline
    \textbf{Actual: Valid}   & 6                & 14,754             \\ \hline
    \end{tabular}
    \caption{Shows the confusion matrix generated by the results from evaluating the ERT model on the testing dataset.}
    \label{tab: ert-confmatrix}
\end{table}

\subsection{Multi-Layer Perceptron}

MLP \cite{haykin1994neural} is a supervised learning algorithm that learns a function $f$ that given an input in space $R^m$ produces an output in space $R^n$ by training on a given dataset. It consists of an input layer, an output layer, and one or more layers in-between called hidden layers. Each layer contains a number of neurons representing parameters of the 
network. Specifically, for the input and output layers those parameters represent the input $(\Vec{x}:x_1,x_2,...,x_m)$ and output values $(\Vec{y}:y_1,y_2,...,y_n)$ of function $f$. The neurons between two contiguous layers $(l_{i-1},l_{i})$ are fully connected with each other through weighted links and the values of the neurons of layer $l_i$ are formed as a weighted linear summation of the neurons on layer $l_{i-1}$ plus some bias $b_i$. Before these values are fed forward from $l_i$ to $l_{i+1}$ a non-linear function, called the activation function, can be applied (e.g. hyberbolic tangent) to introduce non-linearity to the model. 

\begin{figure}[ht]
    \centering
    \includegraphics[width=0.5\textwidth]{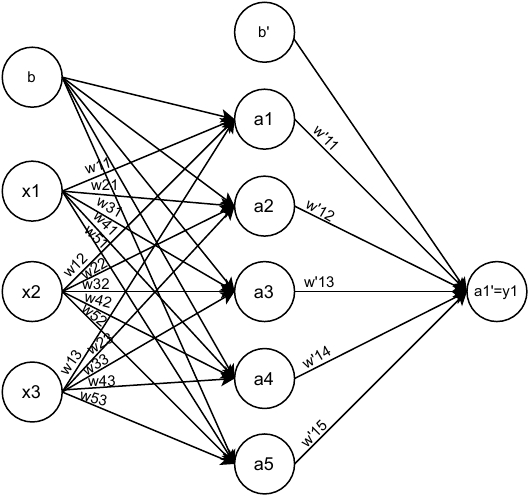}
    \caption{A MLP with a single hidden layer of 5 neurons, 3 inputs and 1 output.}
    \label{fig:mlp_example}
\end{figure}

The number of layers and the number of neurons are referred to as hyperparameters of a neural network which need to be tuned for optimal results. Cross-validation techniques can be used to find the ideal values.
Figure \ref{fig:mlp_example} shows a MLP with a single hidden layer, $R^3$ input and $R^1$ output.
The calculations that take place for the output(s) of each layer is presented below, where $W$ is the matrix of the link weights, $\Vec{x}$ the input vector of the layer, b the bias, $\Vec{z}$ the output before applying an activation function $\phi$ and $\Vec{a}$ the output after applying the activation function:
\[
W\Vec{x}=
\begin{bmatrix}
w_{11} & w_{12} & \cdots & w_{1m}\\
w_{21} & w_{22} & \cdots & w_{2m}\\
\vdots & \vdots & \ddots & \vdots\\
w_{k1} & w_{k2} & \cdots & w_{km}
\end{bmatrix}
\begin{bmatrix}
x_1\\x_2\\ \vdots\\x_m
\end{bmatrix}
+b
=\begin{bmatrix}
z_1\\z_2\\ \vdots\\z_k
\end{bmatrix}
\]
\[
\phi(
\begin{bmatrix}
z_1\\z_2\\ \vdots\\z_k
\end{bmatrix}) 
= \begin{bmatrix}
a_1\\a_2\\ \vdots\\a_k
\end{bmatrix}
\]
Once all layers have received and fed forward their respective values, the output layer produces an output which is evaluated against the ground truth using an error function. Training is then performed by applying backpropagation and optimization. In this stage, an optimization algorithm, like gradient descent, is used to minimize the error function between the predicted and ground truth values. This is achieved by back-propagating the error of the output to the hidden layers and adjusting the values of the weighted links between the neurons. In this way, the next forward pass should produce better results since the weights have been adjusted according to the expected output. By applying multiple iterations of feed-forward and backpropagation, the neural network optimizes its weights to ``learn'' the function $f$ that will produce outputs with minimal error.

\begin{figure}[ht]
    \centering
    \includegraphics[width=0.64\textwidth]{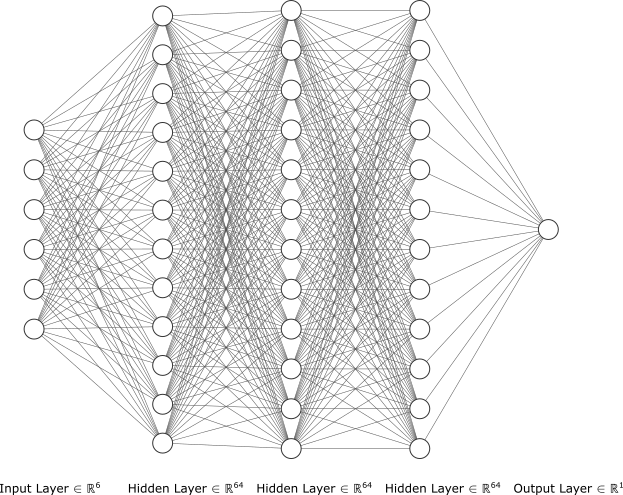} 
    \caption{Architecture of the Multi-Layer Perceptron network.(Generated using \cite{LeNail2019})}
    \label{fig:mlp_arch}
\end{figure}

\paragraph{Architecture}
The MLP used for our application consists of three hidden layers with sixty-four neurons each (see Figure \ref{fig:mlp_arch}). The optimizer chosen was \textit{Adam} \cite{adam-paper}. We used a batch size of thirty-two, and an adaptive learning rate, meaning that the learning rate decreases if there is no training loss decrease in two consecutive epochs. The rest of the parameters were kept as their default values.

\paragraph{Results}
The MLP model achieves 99.95\% accuracy in identifying the valid particle track in a sample (A-det metric). In  38.32\% of the samples where the valid track is identified, there are also some false positives (A-det$|$conf metric). In 92.92\% of the samples, the valid track is given the highest probability of being valid (A-det$|$high metric). Finally, the valid track is not detected in only 0.05\% of samples (A-notdet metric). These results are compiled in Table \ref{tab: mlp-metrics}. Table \ref{tab: mlp-confmatrix} shows the confusion matrix generated by the results from evaluating the model with the testing dataset.

\begin{table}[!h]
    \centering
    \begin{tabular}{|l|l|}
    \hline
    \textbf{Metric}   & \textbf {Result}  \\ \hline
    A-det                & 99.95\% \\ \hline
    A-det$|$conf                & 38.32\%    \\ \hline
    A-det$|$high                & 92.92\%  \\ \hline
    A-notdet                & 0.05\%   \\ \hline
    Time to Train     & 4 hours   \\ \hline
    Time to Predict/sample & 120 $\mu s$ \\ \hline
    \end{tabular}
    \caption{Shows the results for some metrics used to evaluate the model. The MLP model executed on a multi-core CPU.}
    \label{tab: mlp-metrics}
\end{table}

\begin{table}[!h]
    \centering
    \begin{tabular}{|l|l|l|}
    \hline
    & \textbf{Predicted: Invalid} & \textbf{Predicted: Valid} \\ \hline
    \textbf{Actual: Invalid} & 561,994             & 29,469           \\ \hline
    \textbf{Actual: Valid}   & 7                & 14,753             \\ \hline
    \end{tabular}
    \caption{Shows the confusion matrix generated by the results from evaluating the MLP model with the testing dataset.}
    \label{tab: mlp-confmatrix}
\end{table}

\subsection{Convolutional Neural Network}
CNNs are a type of
neural network known to perform better on data where spatial locality is important (e.g. images). A common CNN consists of Convolutional, Pooling and a Fully Connected Layers.  Convolutional layers are the most important layers in a CNN, including an input, a number of kernels, and an output (feature map). 
\begin{itemize}
\item 
The input to a CNN layer is an N-dimensional array (for example, a colored image with height, width and RGB values) given as an input to the network or produced by a previous layer. 
\item
A kernel is a 2-dimensional array of weights, usually 3x3, that is used to apply convolution on the input. The kernel is shifted on the input, based on a stride, and on each shift a dot product is applied between the respective area of the input and the kernel to produce a single element of the output (also known as feature map). The process continues until the whole input has been processed and the whole matrix of the feature map has been produced. The weights of the kernel are the parameters learned by the neural network and are adjusted between iterations through the process of backpropagation and gradient decent.
\item
The result produced by applying convolution on the input using the provided kernel is the output (feature map) of the layer. An activation function is finally applied on the output array (e.g. Rectified Linear Unit or ReLU \cite{RELU2010}) to introduce non-linearity in the model.
\end{itemize}
The size of the output is affected by 3 parameters: the \emph{number of kernels}, the \emph{stride} used during  convolution, and the type of \emph{padding}. 
The \emph{number of kernels} determines the depth of the output (e.g. n kernels would generate n outputs, thus an output of depth n). By increasing the depth of the output, we can extract more features through the extra number of weights to be trained.
As mentioned before, the \emph{stride} determines the step that the kernel moves over the input to apply dot product. A stride larger than one would generate an output of size smaller than the input.
\emph{Padding} can be either ``valid'' or ``same''. A valid padding adds no padding to the input which will result in an output of smaller size for kernels greater than 1x1. This occurs because the kernel will be shifted on the input and apply the dot product fewer times than the size of the input. In other words, when a row or column of the kernel exceeds the size of the input the dot product will be aborted. In same padding, the input is padded by zeros to ensure that the output will be of the same size as the input. 

\begin{figure}[!h]
\centering
\begin{subfigure}{\textwidth}
 \centering
    \includegraphics[width=4.2in]{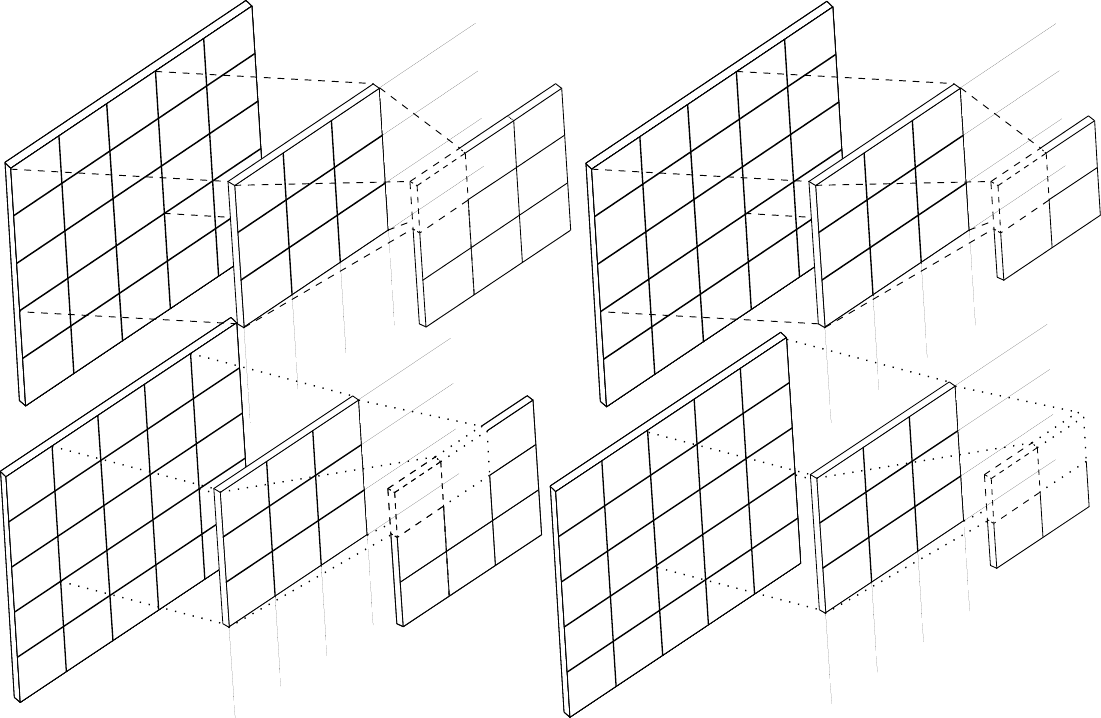}
    \caption{First two steps of convolution with valid padding and a stride of 1 (left) or 2 (right).}
    \label{fig:cnn_valid}
\end{subfigure}
\begin{subfigure}{\textwidth}
 \centering
    \includegraphics[width=4.5in]{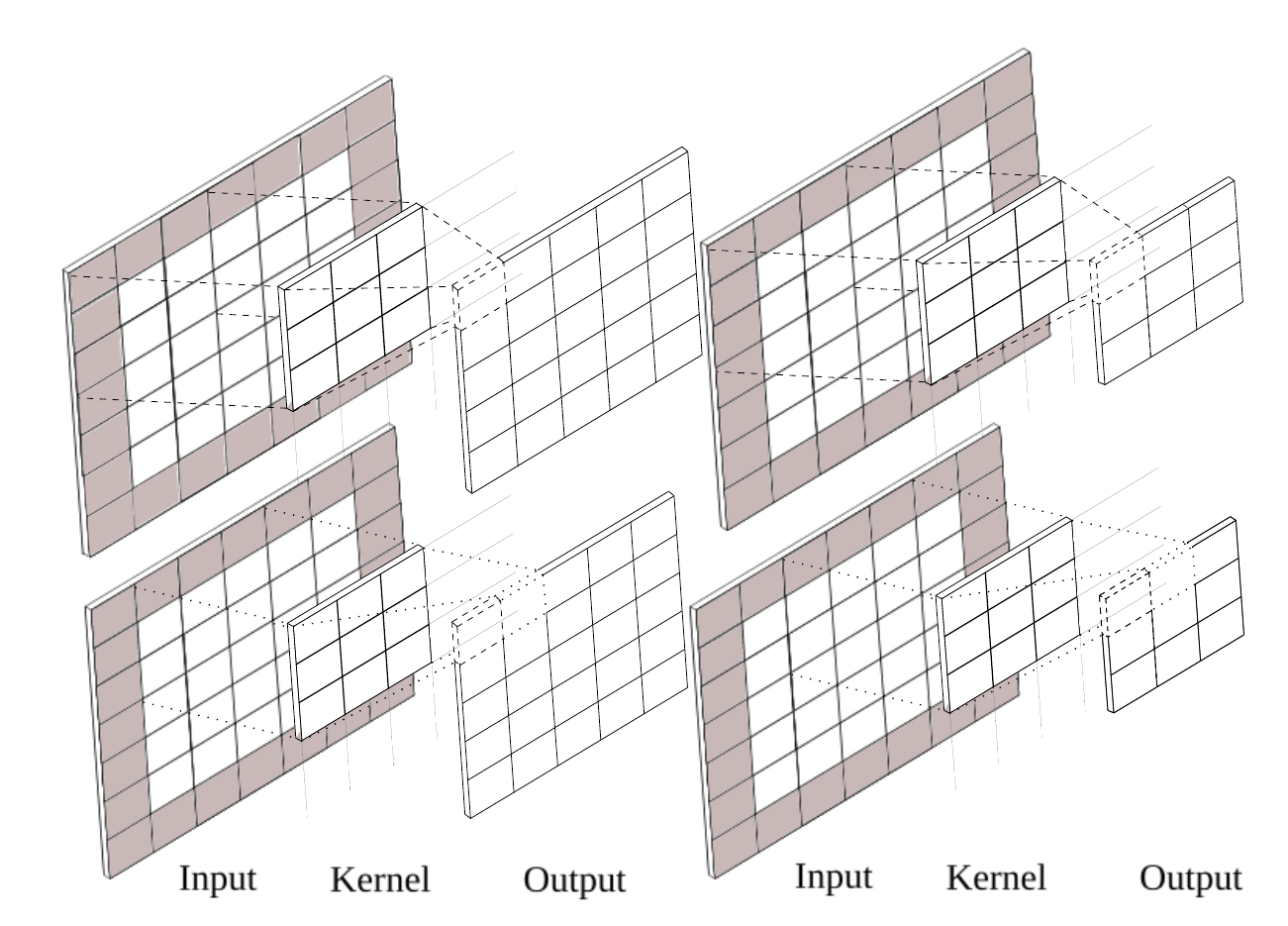}
    \caption{First two steps of convolution with same padding and a stride of 1 (left) or 2 (right). The gray area in the input array represents the padding of zeros added to the input.}
    \label{fig:cnn_same}
\end{subfigure}

\caption{ Examples of convolution between a 5x5 array and a 3x3 kernel, with (a) valid and (b) same padding. The figures show the convolution process with a stride of 1x1 (left column) and a stride of 2x2 (right column). The resulting outputs are generated by applying dot product between the kernel and the respective area of the input, starting from the top left area and shifting the kernel based on the stride. The output for stride 1x1 is generated by shifting the kernel one step at a time (left) while for stride 2x2 by shifting it two steps at a time (right). Figure (a) represents the sizes of the outputs when using valid padding while (b) shows the output for same padding. The sizes of the outputs are affected by both the stride and the padding since those affect how many times the kernel can shift on the input and apply the dot product.}
\label{fig:cnn_example}
\end{figure}

Examples of the convolution process between a 5x5 array and 3x3 kernel for valid and same padding and strides of 1x1 and 2x2 are shown in Figure \ref{fig:cnn_example}. When valid padding is used, (Fig. \ref{fig:cnn_valid}) the kernel can only be applied 3 times horizontally with a stride of 1x1 before moving to the next row. This process is repeated until no more steps are left to be taken both horizontally and vertically for a total of 9 applications of the dot product (3x3 output). When using a stride of 2x2, the kernel is shifted by 2 elements at a time horizontally and 2 elements vertically when a row is completed which allows the application of only 4 dot-products in total (2x2 output). The same process is followed when same padding (fig. \ref{fig:cnn_same}) is applied with the only difference being the introduction of more elements on the edges of the input array to ensure that the output of the convolution (assuming stride 1) will be  the same size as the input. In both cases, the number of kernels equals the depth of the output.

Pooling layers perform downsampling by sweeping a 2-dimensional filter over their input and applying some aggregation function to generate the output. The process is very similar to convolution with the difference being that instead of a dot product, an aggregation function is applied (same as fig. \ref{fig:cnn_valid} but with the kernel only applying an aggregation). The aggregation can either be the maximum  or the average of values in the area under the filter; hence, the names max pooling or average pooling, respectively. This process is used to reduce the amount of parameters to be learned to prevent overfitting and reduce the training time.
The fully connected layer is an  MLP residing at the end of the CNN.

\begin{figure}[!h]
    \begin{center}
        \includegraphics[width=4in]{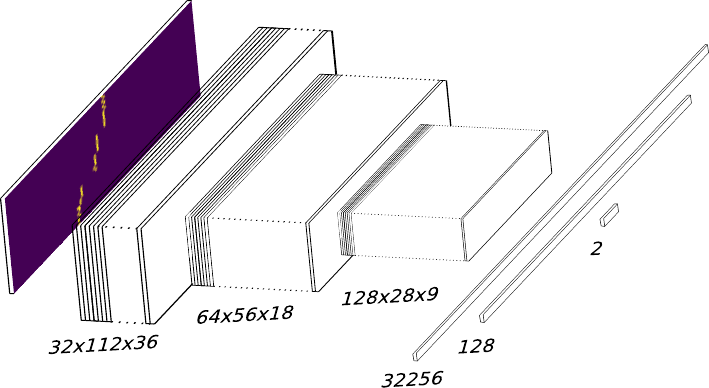} 
    \end{center}
    \caption{Architecture of the CNN network including the feature maps of each convolutional layer and fully connected layer. The number of neurons per layer is also included.}
    \label{fig:cnn_arch}
\end{figure}

\paragraph{Architecture}
The convolutional neural network consists of three convolutional layers of 32, 64 and 128 filters, respectively,  with a 3x3 kernel size, each followed by a 2x2 max-pooling layer with same padding and dropout of 0.25, 0.25 and 0.4 (see Fig. \ref{fig:cnn_arch}). Dropout\cite{dropout} is a technique where a percentage of the available neurons of the network are randomly selected and deactivated during training in an attempt to prevent overfitting. Two dense layers of 128 and two neurons respectively follow the convolutional layers with a droupout of 0.3 between them. The activation function for all layers except the last one is Leaky ReLU\cite{Xu2015EmpiricalEO}, while softmax is used for the last layer. Adam was used as the optimizer and categorical cross-entropy as the loss function. We trained the model for twenty epochs using a batch size of 32.

\paragraph{Results}
 The CNN model achieves good results, with 99.96\% accuracy in identifying the valid particle track in a sample (A-det metric). In approximately 53\% of the samples where the valid track is identified, there are also some false positives (A-det$|$conf metric). In 90.2\% of the samples, the valid track is given the highest probability  (A-det$|$high metric). Finally, the valid track is not detected in 0.04\% of the samples (A-notdet metric). These results are compiled in Table \ref{tab: cnn-metrics}. Table \ref{tab: cnn-confmatrix} shows the confusion matrix generated by the results from evaluating the model with the testing dataset.

\begin{table}[!h]
    \centering
    \begin{tabular}{|l|l|}
    \hline
    \textbf{Metric}   & \textbf {Result}  \\ \hline
    A-det                & 99.96\% \\ \hline
    A-det$|$conf                & 52.99\%    \\ \hline
    A-det$|$high                & 90.22\%  \\ \hline
    A-notdet                & 0.04\%   \\ \hline
    Time to Train     & 8 hours   \\ \hline
    Time to Predict/sample & 1.2 ms \\ \hline
    \end{tabular}
    \caption{Shows the results for some metrics used to evaluate the model. The CNN model executed on one Tesla V100-SXM2-16GB.}
    \label{tab: cnn-metrics}
\end{table}

\begin{table}[!h]
    \centering
    \begin{tabular}{|l|l|l|}
    \hline
    & \textbf{Predicted: Invalid} & \textbf{Predicted: Valid} \\ \hline
    \textbf{Actual: Invalid} & 552,808              & 38,655             \\ \hline
    \textbf{Actual: Valid}   & 6                & 14,754             \\ \hline
    \end{tabular}
    \caption{Shows the confusion matrix generated by the results from evaluating the CNN model with the testing dataset.}
    \label{tab: cnn-confmatrix}
\end{table}


\subsection{Recurrent Neural Network}
Another approach that we investigate is
the use of RNN\cite{rnn-guide}. RNN is a type of neural network that is used to predict
features in datasets that present a sequence. For example, given a sequence of temperature data for
the days in the previous month, an RNN is able to infer the temperatures of the next few days in the
future. RNNs incorporate the connection that exists between sequential data to infer
the next data points in the sequence. To employ them for the needs of identification, we make use of 
the following observation: applying a well trained RNN on a sequence that does not follow the same 
pattern as the training data (i.e. providing a false sequence)  will result in predictions with a
high degree of error.  

In the context of particle trajectories, a track can be presented as a sequence of hit detections on sequential layers of wires. We used this observation to train an RNN using Gated Recurrent Units\cite{cho2014learning} (GRU) layers that given a subset of a track can predict its missing parts. 
Specifically, our model is trained to infer the detections of the last 12 layers given information of the first 24. 
The RNN is trained on the same dataset as the models for trajectory classification, except that only data for valid particle tracks is utilized. This allows the RNN to predict valid particle tracks based on partial previous sensor activation patterns. 
Since the RNN is trained on only valid particle tracks, it will give incorrect predictions for invalid particle tracks. By passing all particle candidates through the RNN, we produce a new set of inferred candidate tracks. 
By measuring the spatial distance of the inferred tracks and the actual track in the dataset, we can extract the tracks that are invalid and those that are valid. An inferred track that has a large distance from the actual candidate track in the dataset is considered invalid, and thus the candidate that was used to generate it is also classified as an invalid. 
On the other hand, when the candidate track in the dataset has a small distance from the inferred one it likely means that the inferred track is valid and so is the track that generated it. 
This process allows us to eliminate most of the invalid tracks and identify possible valid tracks.

\begin{figure}[h]
    \begin{center}
        \includegraphics[width=\textwidth]{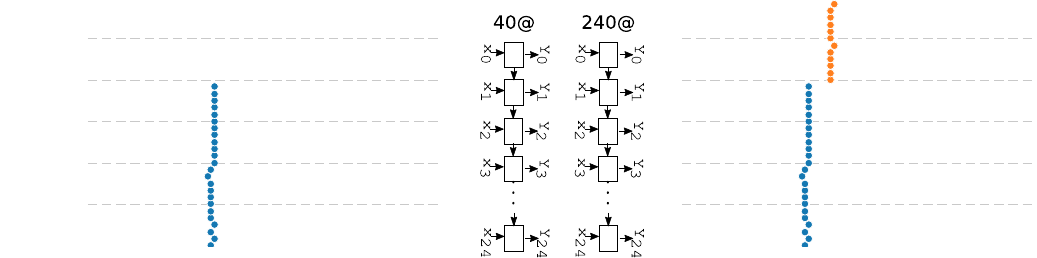} 
    \end{center}
    \caption{Architecture of the recurrent neural network.}
    \label{fig:rnn_arch}
\end{figure}

\paragraph{Architecture}
We use two stacked GRU layers, the first containing 40 units while the second containing 240 units (Figure \ref{fig:rnn_arch}). The input to the model is a vector of features, with each feature occupying one time step. As for the optimizer, we used \textit{RMSprop} \cite{rmsprop-presentation}.

\begin{figure}[H]
    \begin{center}
           \includegraphics[width=6in]{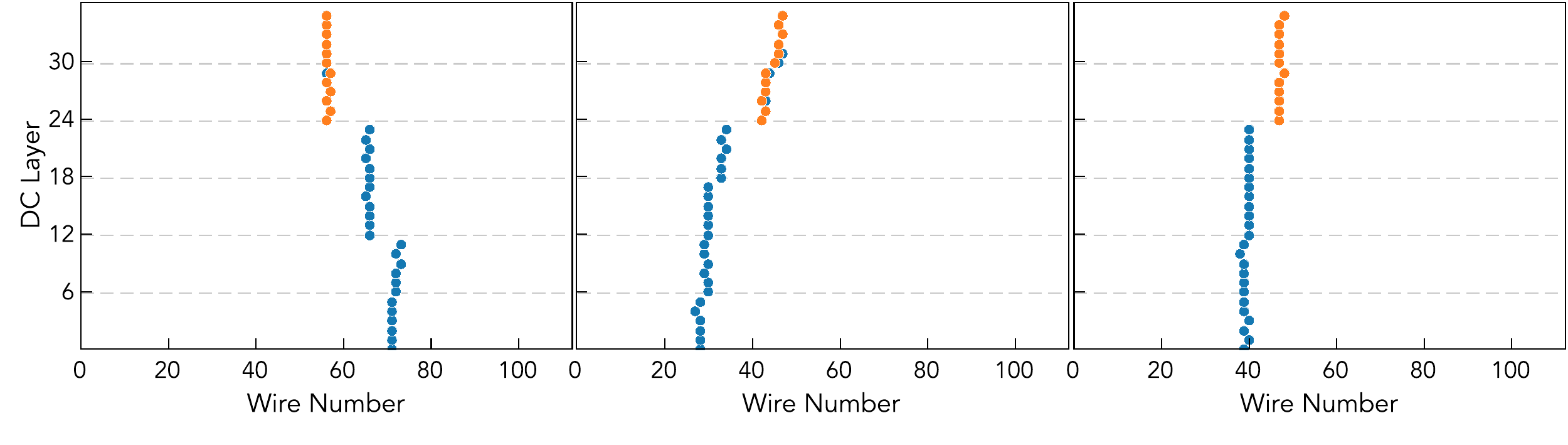} 
    \caption{Shows three separate valid particle tracks (blue) and the predictions of the RNN for part of them (orange). If the RNN correctly predicted the track, then the blue and orange overlap. The small spatial distance between the predicted portion of the tracks and the actual portion of the tracks means that the actual tracks are likely valid.}
    \label{fig: valid-overlapped-trajectories}
    \end{center}
\end{figure}

\begin{figure}[H]
    \begin{center}
        \includegraphics[width=6in]{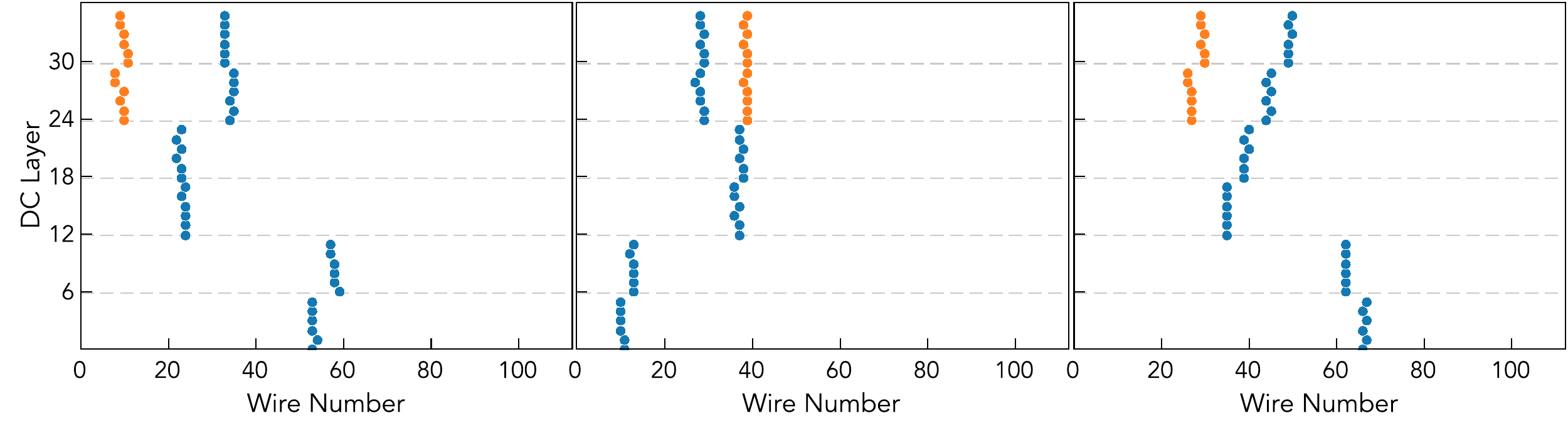} 
    \caption{ Shows three separate invalid particle tracks (blue) and the prediction of the RNN for part of them (orange). The larger spatial distance between the predicted portions of the tracks and the actual portions of the tracks means that the actual tracks are likely invalid.}
    \label{fig: invalid-overlapped-trajectories}
    \end{center}
\end{figure}

\paragraph{Results}
The RNN achieves an average mean absolute error (MAE) of about 1.57. In the context of this problem, this means that the predicted particle trajectories of the model were on average about 1.57 sensors away from the actual particle trajectories. Figures \ref{fig: valid-overlapped-trajectories} and \ref{fig: invalid-overlapped-trajectories} show examples of actual particle tracks overlapped with the predicted portions of the particle tracks from the RNN. For the classification process we used Mean Absolute Error (MAE) as the metric for spatial distance and explored with different values as the maximum distance. 

\begin{table}[h]
    \centering
    \begin{tabular}{|c|c|c|c|}
    \hline
    \textbf{Metric} & \textbf{Max Dist: 2} & \textbf{Max Dist: 3} & \textbf{Max Dist: 4}\\ \hline
    \textbf{A-det (\%)} & 91.98              & 97.58               & 99.0 \\ \hline
    \textbf{A-det$|$conf (\%)} & 69.65              & 83.32               & 90.65 \\ \hline 
    \textbf{A-det$|$high (\%)} & 71.74              & 73.91               & 74.17 \\ \hline
    \textbf{A-notdet (\%)} & 8.02               & 2.41                & 1.0 \\ \hline
    \end{tabular}
    \caption{ Shows the results of the metrics used for the evaluation of the RNN model for different maximum distances.}
    \label{tab: rnn-metrics}
\end{table}
\begin{table}[h]
    \centering
    \begin{tabular}{|c|c|c|c|}
    \hline
    & \textbf{Max Dist: 2} & \textbf{Max Dist: 3} & \textbf{Max Dist: 4}\\ \hline
    \textbf{True Positive (\%)} & 91.98             & 97.58             &    99.0 \\ \hline
    \textbf{False Negative (\%)} &    8.02              &      2.41      &   1.0 \\ \hline 
    \textbf{True Negative (\%)} &   92.45               &   88.35         &    84.0 \\ \hline
    \textbf{False Positive (\%)}   & 7.55                &  11.65            &   16.0 \\ \hline
    \end{tabular}
    \caption{Shows the confusion matrix values generated by the results from evaluating the RNN model with the testing dataset. The true positive  and false negative percentages are presented as a fraction of the valid tracks in the dataset. Accordingly, the percentage of true and false negative are presented as a fraction of the invalid tracks.}
    \label{tab: rnn-confmatrix}
\end{table}

Tables \ref{tab: rnn-metrics} \& \ref{tab: rnn-confmatrix} show the performance of the RNN for the different maximum distances.  As expected, a larger maximum distance helps to identify more valid tracks but increases the amount of invalid tracks misclassified as valid. On the other hand, a smaller distance decreases the amount of tracks falsely identified as valid but also increases the amount of valid tracks that are missed by the algorithm.  Since there are no probabilities infered from the network to classify as valid or invalid, the value we used to compute the A-det$|$conf metric is the distance between the inferred and the actual track candidate (the smaller the better). The performance of the RNN shows that this approach can be promising, however, the three other machine learning networks outperform it.

\section{Performance Summary}
We have presented four machine learning models applied to perform track classification for CLAS12 drift chambers along with a brief introduction for each method. In this section, we summarize the results of all four methods in tables \ref{tab: trajectory-classification-metrics}, \ref{tab:sum_conf_matrix} below: 
\begin{table}[!htb]
    \centering
    \small\addtolength{\tabcolsep}{-2.2pt}
    \begin{tabular}{|c|c|c|c|c|c|c|}
        \hline
        \textbf{Model} & \textbf{A-det(\%)} & \textbf{A-det$|$conf(\%)} & \textbf{A-det$|$high(\%)} & \textbf{A-notdet(\%)} & \textbf{Training} & \textbf{Inference} \\
        \textbf{Type} & & & &  & \textbf{Time} & \textbf{Time/sample}\\ \hline
        ERT & 99.96 & 40.72 & 97.09 & 0.04   & 506 sec & 306 $\mu s$  \\ \hline
        MLP & 99.95 & 38.92 & 92.92 & 0.04   & 4 hours & 120 $\mu s$ \\ \hline
        CNN & 99.96 & 52.99 & 90.22 & 0.04   & 8 hours & 1.2 $ms$ \\ \hline
        RNN & 99.0  & 90.65 & 74.17 & 1.0    & 3 hours & 343 $\mu s$ \\ \hline
    \end{tabular}
    \caption{Summarizes the results for the four models for particle trajectory classification. The MLP and ERT models executed on a multi-core CPU, while the CNN executed on one Tesla V100-SXM2-16GB.}
    \label{tab: trajectory-classification-metrics}
\end{table}

\begin{table}[!htb]
    \centering
    \small\addtolength{\tabcolsep}{-2.2pt}
    \begin{tabular}{|c|c|c|c|c|}
        \hline
        \textbf{Model} & \textbf{True Positive} & \textbf{False Negative} & \textbf{True Negative} & \textbf{False Positive} \\ 
        \textbf{Type} & \textbf{(\%)}& \textbf{(\%)}& \textbf{(\%)}& \textbf{(\%)}\\ \hline
        ERT & 99.69 & 0.04 & 94.68 & 5.31 \\ \hline
        MLP & 99.95 & 0.05 & 95.02 & 4.98  \\ \hline
        CNN & 99.96 & 0.04 & 93.46 & 6.53  \\ \hline
        RNN & 99.0  & 1.0  & 84.0  & 16.0 \\ \hline
    \end{tabular}
    \caption{Summarizes the confusion matrix values generated by the results from evaluating each model with the testing dataset. The true positive  and false negative percentages are presented as a fraction of the valid tracks in the dataset. Accordingly, the percentage of true and false negative are presented as a fraction of the invalid tracks.}
    \label{tab:sum_conf_matrix}
\end{table}

\begin{figure}[!hb]
    \centering
    \includegraphics[width=2.8in]{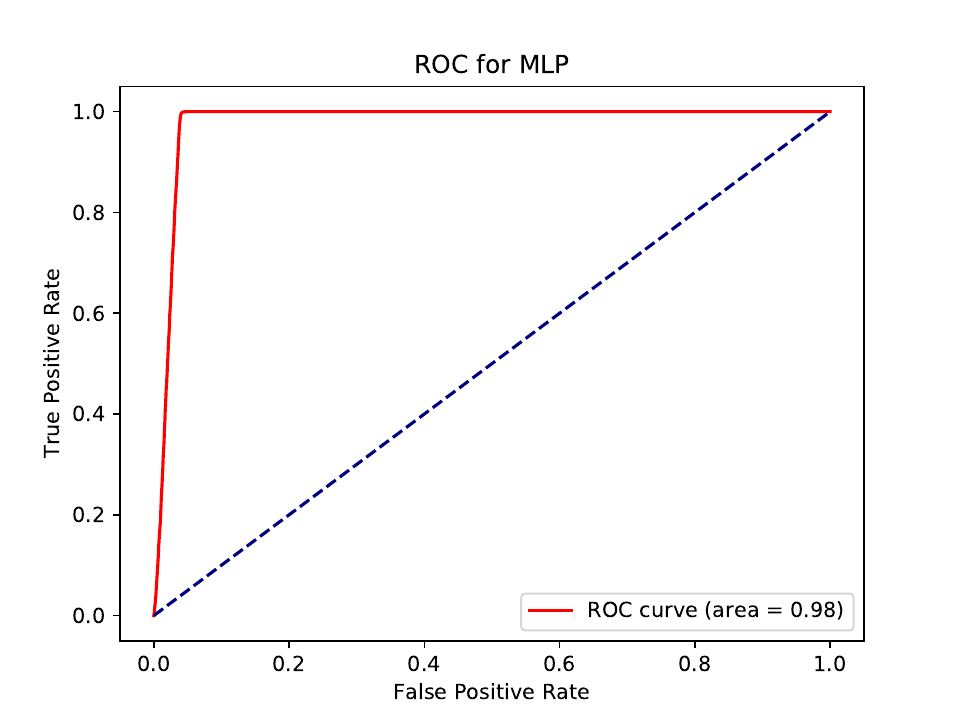}
    \includegraphics[width=2.8in]{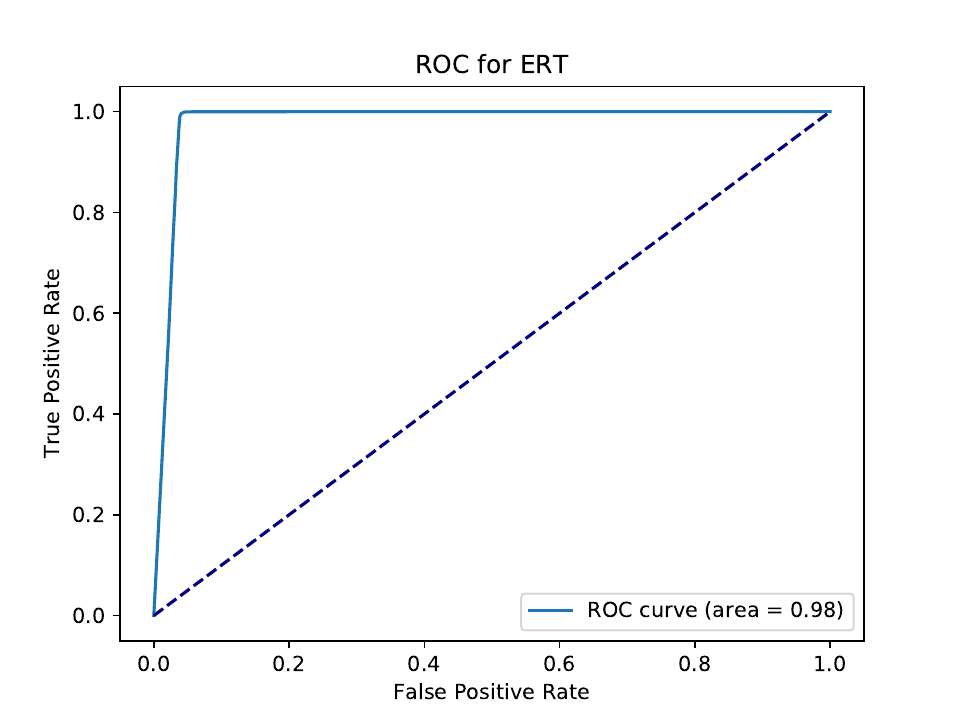}
    \includegraphics[width=2.8in]{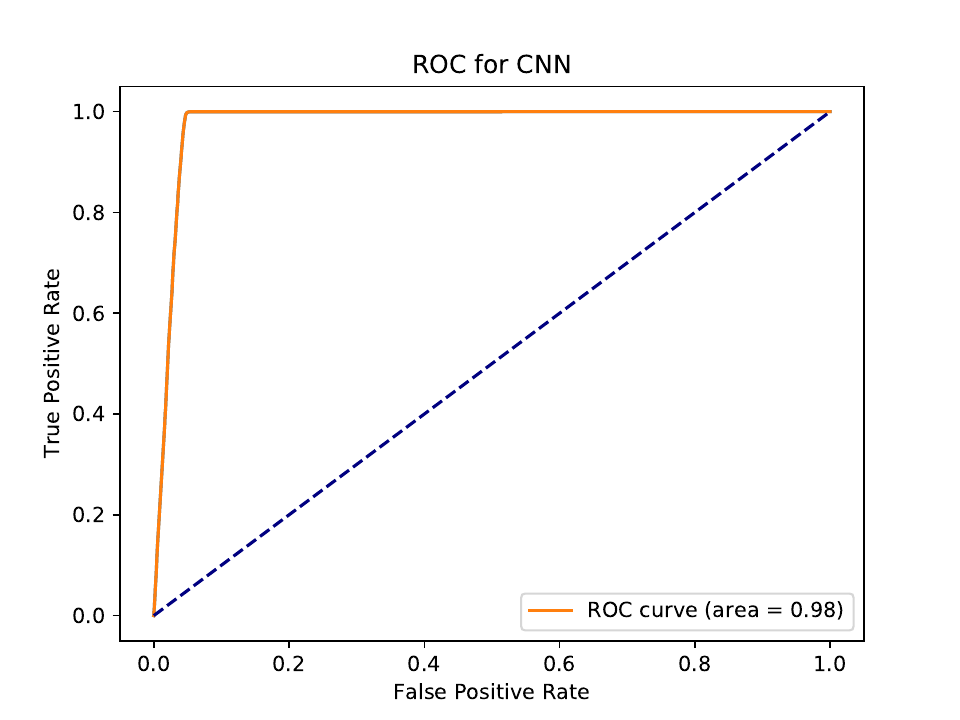}
    \caption{ROC curves produced from the MLP (left), CNN (middle), and ERT (right) models.}
    \label{fig:roc_curves}
\end{figure}

We can see that all methods manage to identify the valid track very accurately with over 99\% success. ERT performs the best in terms of accuracy, while MLP comes close second, however, it performs 2.5 times better in terms of inference speed. CNN gets the third place  in both accuracy and inference time; an important factor for that being that it utilizes a much larger data format (4032 features vs 6 vs 36). The RNN falls quite behind in terms of accuracy, misclassifying about 1\% of the valid tracks versus 0.04\% of the rest of the models. Moreover, the percentage of events that contain false classifications amount to 90.65\%, more than double of what the ERT and MLP models achieved. Out of this study we decided to implement the MLP algorithm in the track reconstruction process of CLAS12 due to its high speed and high accuracy in detecting the valid track. Figure~\ref{fig:roc_curves} presents the ROC curves of the three models that output probabilities (MLP, CNN and ERT) where we can see that their performance is almost identical.

\section{Software Implementation}
\label{sec:soft_impl}
To conduct this study, we implemented a framework
that simplifies the process of training, validation, and prediction.
Our software\footnote{Code and data available at: \url{https://github.com/gavalian/clas12ai/tree/master/ml}} is written in the Python programming language to accelerate the development
process and make the software easily maintainable, reusable, and extensible. We use TensorFlow 2 \cite{tensorflow} for the creation, training, and validation of the CNN and RNN machine learning models while scikit-learn~\cite{scikitlearn} was used for the ERT and MLP models.

Since our datasets are sparse, we use the svmlight data format \cite{svmlight} which is
designed for labeled data with sparse features. To load and store datasets, we make use of the scikit-learn library. Scikit-learn is also utilized to generate validation datasets by randomly picking data entries from the training dataset. To enable easy distribution of the software, we provide a YAML\cite{yaml} file that contains the required dependencies that can be provided to software like Anaconda~\cite{anaconda} to handle all software dependencies.

Our software consists of two separate operations, \texttt{train}
and \texttt{test}. Each operation is implemented
as a subcommand that accepts its own parameters (similar to how git 
commands work: \texttt{git commit $\langle$args$\rangle$}, 
\texttt{git branch $\langle$args$\rangle$}, etc.). \\
The \texttt{train} subcommand, as shown from its name, is used to train a 
new model. It takes a set of required and optional arguments; the required ones include the dataset to use for training, the format that the dataset has (4032, 36 or 6 features), the model architecture to be used (CNN, MLP, ERT, RNN), and the path to store the trained model.
Optionally, the user can also set the number of 
training epochs to go through, the batch size, and a separate validation file.

The \texttt{test} subcommand performs the respective evaluation operation;
accepted arguments include the trained model to use, the dataset
to test it on, the directory to store the performance results (all required), and the batch size (optional). When testing completes, it
generates a report for the user with the accuracy results, confusion matrix
etc.

Extending the framework with new models is simple. A new model 
architecture can be implemented by subclassing a built-in abstract class
and implementing two methods; namely the \texttt{build()}, and
\texttt{preprocess()} methods. The \texttt{build()} method is where the
architecture of the new model is described using Keras notations,
compiled, and returned to the framework. The \texttt{preprocess()} method
is provided so that the user can perform any preprocessing that
might be needed for the specific model, application, and dataset. For example, to test the CNN model the input needs first to be transformed to a
36x112x1 tensor to fit the CNN's expected input format ($ width \times height \times channels$).

\section{CLAS12 Tracking}

In our studies, we used only one track type, namely negatively charged tracks as training and validation datasets.
However, the final implementation of the MLP had 6 input features, representing the mean values of clusters in each super-layer, for both positively and negatively charged  track candidates  . Thus,
classification in the output results into one of three possible cases, namely: ``false track'', ``negatively charged track'', and ``positively charged track''. 
For training, we use events where two tracks were identified in one of the sectors of the drift chambers and use valid track features as positive samples
for the neural network. As negative sample (``false track''), two candidates are constructed by swapping clusters of two tracks. A random number of clusters (one or two) are swapped in each event (determined by a random number generator). This method assures that no valid reconstructed track can be represented as a ``false track'' in the training sample. It also provides better accuracy,  because it produces ``false'' track candidates that are similar to the real track by swapping only one or two clusters.

The trained network is, then, used to identify track candidates from real experimental data, and predictions are compared 
with the tracks reconstructed  by  the conventional algorithm. 
In this case, we do not use cuts/thresholds for classification. Classification is done by picking the track candidate with the highest probability. From our tests, most of the time ($> 98.5\%$), the valid track has a probability of $>0.85$, but if so happens that the highest probability is 0.3, we pick the respective track and pass it to the tracking algorithm to consider. 

\begin{figure}[!hb]
\begin{center}
\includegraphics[width=6.0in]{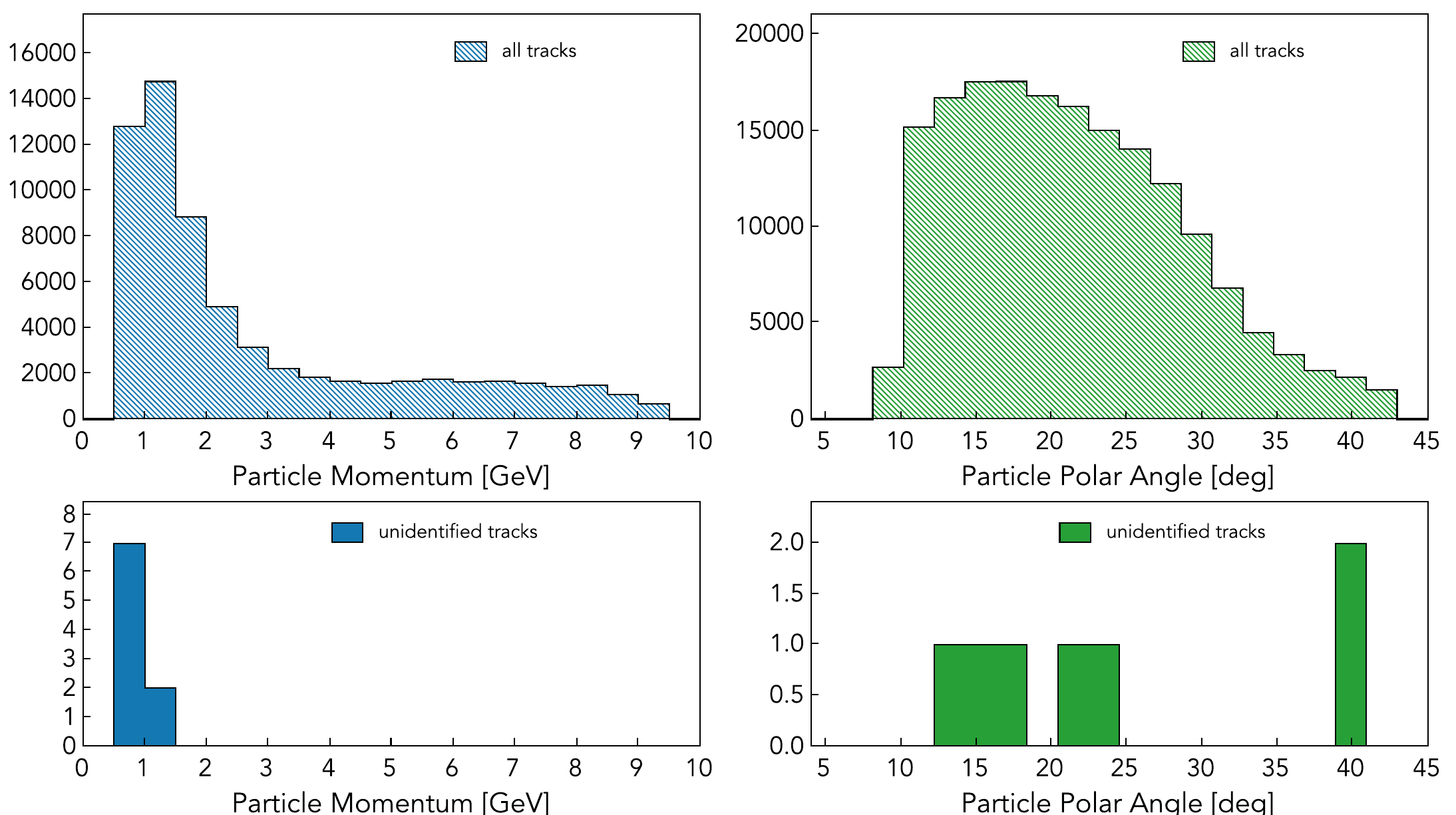}
\caption {The distribution of tracks as a function of particle momentum and polar angle reconstructed by conventional tracking algorithm (top row).
Distribution of tracks as a function of momentum and angle for track candidates that were not identified by ML (bottom row). }
 \label{ai:distribution}
 \end{center}
\end{figure}

\begin{figure}[h]
\begin{center}
\includegraphics[width=6.0in]{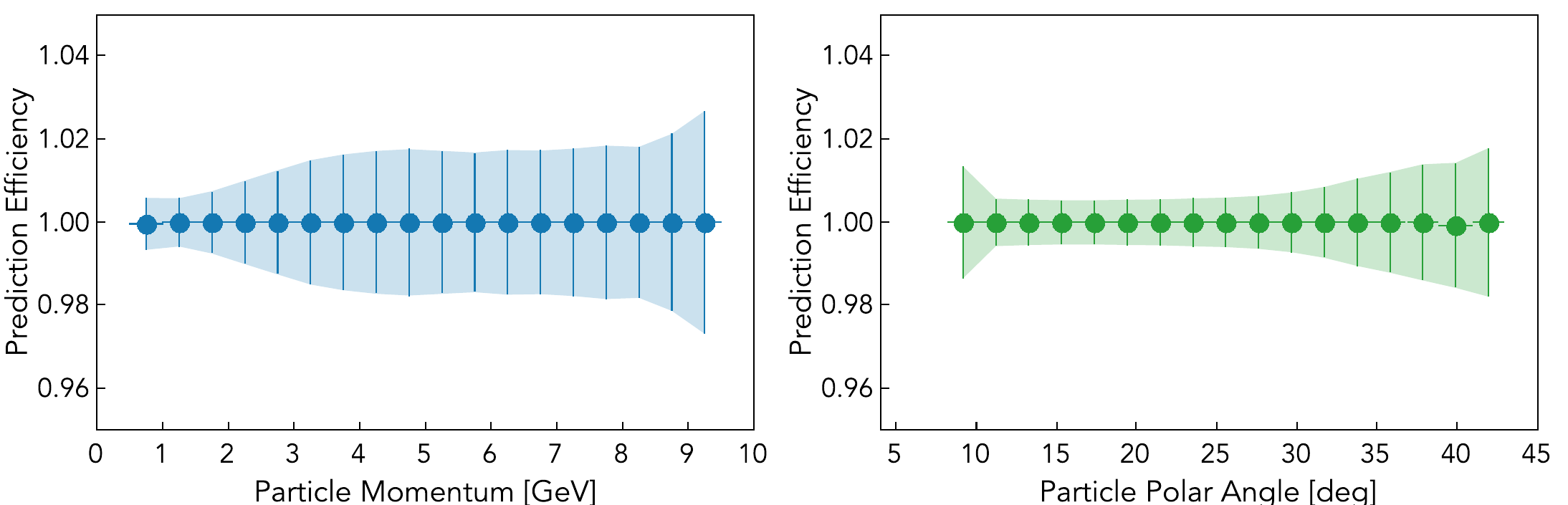}
\caption {Efficiency of track identification using suggestions from MLP network as a function of particle momentum (left) and polar angle (right). }
 \label{ai::efficiency}
 \end{center}
\end{figure}

The distribution of negative tracks is shown in Figure~\ref{ai:distribution} (top row) as a function of particle momentum and polar angle in a laboratory reference frame. In Figure~\ref{ai:distribution} (bottom row), the distribution of tracks that are not 
identified by the network as good tracks is shown as a function of momentum and angle. 
Only a tiny fraction (9 out of 65224 tracks) was not identified by the neural network as a good track candidate. In Figure~\ref{ai::efficiency}, the efficiency of track identification is shown as a function of particle momentum and polar angle. The track identification 
accuracy for both positively and negatively charged particles is summarized in Table~\ref{tab:efficiency}, showing efficiency $>99.9\%$ for both particle charges.

\begin{table}[!ht]
\begin{center}
\begin{tabular}{|l|r|r|r|r|}
\hline
Particle Charge & Conventional Tracks & ML Predicted &  ML missed & Efficiency \\
\hline
\hline
negative & 65224 & 65215 & 9 & 0.999862 \\
positive & 177434 & 177411 & 23 & 0.999865 \\
\hline
\end{tabular}
\end{center}
\caption{Summary of track identification efficiency negatively and positively charged particles.}
 \label{tab:efficiency}
\end{table}

The implementation of our MLP network was ported into the standard CLAS12 software. The ML module follows the clustering algorithm and composes all
possible combinations of tracks found from each segment. All track candidates are, then, evaluated by the ML module. The suggestions 
of tracks are saved into intermediary data structures and are passed to the tracking algorithm, which uses these suggestions to construct and fit the
final tracks.
Our study with experimental data shows that our neural network can identify good tracks with efficiency above $99.9\%$ and provide a $35\%$ speedup compared to the existing tracking code. The time reduction comes from the fact that much fewer invalid tracks need to be fitted with the Kalman-Filter, tracks which would be thrown away later due to non-convergence. Fitting an event's track candidates takes about 350-400 ms (depending on combinatorics), thus, ML track candidate finding essentially removes this overhead ($120 \mu s \, vs\, 350ms$) by minimizing the number of tracks that conventional algorithms have to consider and intensively fit.

\section{Conclusion and Future Work}

We have demonstrated the performance of four machine learning models, aiding the CLAS12 reconstruction code with track classification.
We have shown that all four models ERT, MLP, RNN, and CNN perform well for this task. The MLP shows the best results in terms of accuracy
and inference speed on real data. A small systematic study was implemented, showing that inference accuracy is increased substantially by training the models on datasets with false samples very similar to true ones. Finally, we evaluated the performance  of the MLP model in comparison to the traditional track candidate selection algorithm method, which showed an equally good track identification efficiency and resulted in $35\%$ track reconstruction speed up. 
In the future, we plan to extend the ML model by introducing a preprocessing step to further filter the candidate tracks. An RNN will be introduced to give an estimate of the path the true track will follow. This estimation will be used to narrow down the number of candidates the MLP needs to consider.
This network will be first implemented in the CLAS12 reconstruction software to test improvements in track reconstruction efficiency. Then, it will be complemented with a de-noising auto-encoder developed for CLAS12 tracking \cite{Thomadakis:2022zcd}, which already shows significant improvements for segment identification in high background (high luminosity) conditions. 

\section{Acknowledgments}

This material is based upon work supported by the U.S. Department of Energy, Office of Science, Office of Nuclear Physics under contract DE-AC05-06OR23177, and
 NSF grant no. CCF-1439079 and the Richard T. Cheng Endowment. The authors would like to thank Raffaella De Vita for helping in processing data with CLAS12 reconstruction software and the anonymous reviewers for their constructive comments. This  work  was  performed  using  the  Turing  and  Wahab computing clusters at Old Dominion University (ODU).
 
\newpage
\bibliography{references}
\bibliographystyle{ieeetr}

\end{document}